# Navigating Spatial Inequities in Freight Truck Crash Severity via Counterfactual Inference in Los Angeles

## Author information

**Yichen Wang**
Ph.D. candidate
The Research Center for Digital Production and Smart Logistics, Tsinghua University
Department of Logistics and Transportation at Tsinghua University
Email: wang-yc22@mails.tsinghua.edu.cn

**Hao Yin**
Ph.D. candidate
The Research Center for Digital Production and Smart Logistics, Tsinghua University
Department of Logistics and Transportation at Tsinghua University
Email: yinh24@mails.tsinghua.edu.cn

**Yifan Yang**
Ph.D. candidate
Geospatial Exploration and Resolution (GEAR) Lab
Department of Geography, College of Arts & Sciences, Texas A&M University
Email: yyang295@tamu.edu

**Chenyang Zhao**
Ph.D. candidate
Artificial General Intelligence Lab, University of California, Los Angeles
Computer Science department at University of California, Los Angeles
Email: zhaochenyang@cs.ucla.edu

**Siqin Wang**
Associate Professor
Spatial Sciences Institute
Dornsife College of Letters, Arts and Sciences, University of Southern California
Email: siqinwan@usc.edu

**Abstract:** Freight truck-related crashes pose significant challenges, leading to substantial economic losses, injuries, and fatalities, with pronounced spatial disparities across different regions. This study adopts a transport geography perspective to examine spatial justice concerns by employing deep counterfactual inference models to analyze how socioeconomic disparities, road infrastructure, and environmental conditions influence the geographical distribution and severity of freight truck crashes. By integrating road network datasets, socioeconomic attributes, and crash records from the Los Angeles metropolitan area, this research provides a nuanced spatial analysis of how different communities are disproportionately impacted. The results reveal significant spatial disparities in crash severity across areas with varying population densities, income levels, and minority populations, highlighting the pivotal role of infrastructural and environmental improvements in mitigating these disparities. The findings offer insights into targeted, location-specific policy interventions, suggesting enhancements in road infrastructure, lighting, and traffic control systems, particularly in low-income and minority-concentrated areas. This research contributes to the literature on transport geography and spatial equity by providing data-driven insights into effective measures for reducing spatial injustices associated with freight truck-related crashes.

# 1 Introduction

The demand for road freight transportation has increased significantly in the past few decades due to globalization and the rise of e-commerce, leading to a higher concentration of freight movement in specific geographical areas and corridors (Jain *et al.*, 2020; H. Wang *et al.*, 2021; Yu *et al.*, 2024; Yuan *et al.*, 2023). The spatial distribution of freight activities has resulted in varying levels of exposure to freight truck-involved crashes across different communities, increasing the likelihood of such incidents in certain regions. The Federal Motor Carrier Safety Administration (FMCSA) noted that from 2009 to 2019, the number of large trucks involved in fatal crashes increased by 47%, reflecting not only the increased presence of trucks on the road but also the geographical disparities in associated risks (FMCSA, 2021). Crashes involving freight trucks are more severe than those involving smaller vehicles. Due to their size and weight, freight trucks can cause more significant damage, leading to higher rates of fatalities and serious injuries. In 2020, large trucks were involved in 9% of all fatal crashes, while they only account for 4% of all registered vehicles and 7% of total vehicle miles traveled according to the National Highway Traffic Safety Administration report (NHTSA, 2021). These statistics highlight the importance of understanding the spatial patterns of freight truck-involved crashes and their implications for different regions and communities. Freight truck-involved crashes can cause significant economic losses due to delays, damage to goods, and increased transportation costs (Fountas *et al.*, 2020; Hossain *et al.*, 2023), which can disproportionately affect certain geographic areas.

Given the rising severity of freight truck-involved crashes, numerous studies have been conducted to analyze the influencing factors related to the temporal and spatial distribution of these accidents, as well as the severity of injuries sustained based on the different weather conditions and road types (Ahmed *et al.*, 2018; Yang *et al.*, 2021b; Yu *et al.*, 2024). By examining the spatiotemporal distributions of freight truck-involved crashes, a series of prevention strategies have been proposed and implemented based on the identified risk factors to enhance overall road safety not only from management strategies, such as zone-based restrictions during peak hours (Sze & Song, 2019), etc. but also from the traffic design (Bassan, 2016) and urban planning (Dumbaugh & Rae, 2009; Yu *et al.*, 2024) perspectives. However, these strategies often overlook the spatial justice implications, where certain communities, particularly low-income or minority populations (Najaf *et al.*, 2018), may bear a disproportionate burden of freight-related risks due to their geographical location near major freight corridors or industrial zones. For example, Park and Park (2022) have indicated that roads with fault, snowy, wet surface conditions, and construction areas significantly contribute to crashes by employing multilevel mixed-effects models. Yu *et al.* (2024) have answered how to improve freight truck driving safety performance on urban non-highway roads and spatial areas with highway access. Despite these contributions, there remains a gap in understanding how geographical disparities and spatial inequities influence the distribution and severity of freight truck-involved crashes (Faisal Habib *et al.*, 2024; Yang *et al.*, 2021a). The analysis and prevention of freight truck crashes require empirical research based on solid evidence (McDonald *et al.*, 2019). Hence, most existing studies have only been able to identify factors associated with the spatial distribution of these crashes. Few have addressed the question of which specific factors contribute to the occurrence of a freight truck crash to what extent, and how these factors interact within different geographical contexts, creating a gap between understanding the causes of crashes and developing effective, geographically targeted approaches to mitigate their occurrence and reduce consequences.

While recent studies have put increasing emphasis and begun to explore the causality of traffic crashes (Elvik, 2024; Shinar & Hauer, 2024; Ulak & Ozguven, 2024), studies specifically focusing on freight truck accidents from a transport geography perspective are scarce. This lack of targeted research makes it challenging to understand causative factors unique to freight truck-related crashes, which are often more complex due to the trucks' route choice behaviors and complicated interactions with built environments such as road network patterns and land use patterns. Moreover, these studies have not adequately addressed the spatial justice concerns and counterfactual scenarios, which ask whether a crash would still have occurred if certain conditions had been different in specific geographical contexts. Without this understanding, it is difficult to evaluate the potential impact of proposed interventions or changes in conditions on different regions or communities. Counterfactual analysis is crucial for testing hypotheses about causation and for designing effective prevention strategies that can be adapted to real-world conditions.

To address the research gap in understanding the causal factors influencing freight truck-involved crashes and their spatial distribution, we propose a deep counterfactual inference model from a transport geography standpoint, which offers a more precise analysis of the causal relationships and the potential impact of different interventions on crash severity across different geographical areas. This model allows for the inference of how changes in road conditions, driver behaviors, and environmental factors could affect crash outcomes in various spatial contexts. By simulating counterfactual scenarios, our model enables a detailed comparison between the observed crash data and hypothetical interventions, providing insight into the effectiveness of specific safety measures targeted at specific regions. The study leverages empirical data from the Los Angeles Metropolitan area to evaluate cross-sectoral strategies, focusing on interventions such as improved traffic management systems and road infrastructure enhancements in areas identified as high-risk zones due to spatial injustice factors. This approach enables us to quantify the effectiveness of interventions like lighting and road control device optimization in reducing the severity of freight truck crashes and addressing geographical disparities.

This research contributes to both theoretical and practical domains. Theoretically, we introduce a deep counterfactual inference model that enhances the capacity to analyze the causal factors underlying freight truck crashes within the framework of transport geography and spatial justice. This model represents a substantial advancement in traffic safety research by providing a robust framework for evaluating how changes in various conditions could influence crash severity across different geographical locations. Practically, by simulating the effects of different policy interventions, the model allows us to develop targeted strategies aimed at reducing crash severity and mitigating spatial injustices. This approach offers a clearer understanding of how adjustments in road infrastructure, traffic regulations, and driver behavior can mitigate the dangers associated with freight truck operations, especially in communities disproportionately affected, ultimately improving road safety and reducing crash severity from a spatial equity perspective.

## 2 Literature Review

### 2.1 What factors determine the spatial distribution of freight truck-involved crashes?

Multiple interrelated factors shape the spatial distribution of freight truck-involved crashes, which can be concluded into four primary categories according to the existing literature: road infrastructure, traffic status, weather conditions, and socioeconomic and demographic factors (Hong *et al.*, 2019; Taylor *et al.*, 2018; Yu *et* *al.*, 2024). Road infrastructure plays a critical role in determining where freight truck crashes are likely to occur (Haddad *et al.*, 2023; Ramiani & Shirazian, 2020). The design and quality of infrastructure can either mitigate or exacerbate the risk of crashes. Research has shown that road design and patterns significantly influence crash rates. For instance, roads with complex intersections, sharp curves, and inadequate lane widths create hazardous conditions for freight trucks, which require more space and time to drive (Bassan, 2016). Poorly designed roadways increase the likelihood of crashes by limiting truck drivers' ability to respond to sudden changes in traffic conditions. Road network pattern factors were reported in the work of pedestrian-related crashes by Guo *et al.* (2017) first. In this regard, a pioneer research by Yu *et al.* (2024) has indicated that the regional road network's centrality index should take at least 17.11% responsibility for the spatial patterns of freight truck-involved crashes in the Los Angeles metropolitan area. The presence of dedicated freight lanes is another crucial factor in this category. These lanes help reduce interactions between trucks and smaller vehicles and pedestrians, thereby minimizing potential conflicts and enhancing overall safety. Segregated lanes for trucks lead to fewer collisions by allowing freight vehicles to maintain consistent speeds and reducing the need for frequent lane changes (Shin, 2024a; J. Wang *et al.*, 2020). These findings have emphasized that investments in road design, maintenance, and dedicated infrastructure for freight transportation can reduce crashes, highlighting the importance of infrastructure planning in traffic safety strategies (Yuan & Wang, 2021). However, these studies also underscore several limitations and challenges. While dedicated lanes and improved infrastructure can enhance safety, the implementation of such measures often faces practical constraints, including limited urban space and high construction costs. Furthermore, the effectiveness of these strategies is influenced by other factors such as driver behavior (Mehdizadeh *et al.*, 2019; Newnam *et al.*, 2018; Rashmi & Marisamynathan, 2023), which are not solely addressable through road infrastructures.

Traffic conditions, particularly volume and congestion, are pivotal in influencing the spatial distribution of freight truck crashes. High traffic volumes often correlate with increased crash risks, as they can lead to congestion (Rahimi *et al.*, 2019), which exacerbates the likelihood of accidents. C. Wang *et al.* (2022) have demonstrated that congested conditions not only increase the frequency of crashes but also amplify their severity. In heavy traffic, trucks have limited maneuverability, making it difficult for drivers to navigate safely and avoid collisions. Speed variability is another critical factor (Madarshahian *et al.*, 2024). In regions with mixed traffic flows and inconsistent speed limits, the risk of crashes increases (Duvvuri *et al.*, 2022). Choudhary *et al.* (2018) have shown that regions with frequent speed changes and stop-and-go traffic tend to experience higher rates of truck-involved crashes. This variability creates dangerous hazards and situations where trucks may struggle to maintain safe distances from other vehicles, leading to collisions (Al-Bdairi & Hernandez, 2020; Song & Fan, 2021).

Weather conditions are a well-documented factor affecting the spatial distribution of freight truck crashes (Becker *et al.*, 2022; Kim *et al.*, 2024; Li *et al.*, 2020; Naik *et al.*, 2016). An 18-year longitudinal analysis from 1994 to 2012 in the whole U.S. has concluded that adverse weather, such as rain, snow, and fog, reduces visibility and road traction, significantly increasing crash likelihood (Saha *et al.*, 2016). Park and Park (2022) employed multilevel mixed-effects models to demonstrate that adverse weather conditions contribute substantially to the occurrence of freight truck crashes. These conditions can lead to slippery roads and impaired driver visibility, making it more challenging for truck drivers to control their vehicles and react to sudden changes in traffic conditions. Seasonal variability also plays a vital role, with winter months often experiencing higher crash rates due to snow and ice (Okafor *et al.*, 2022; Wei *et al.*, 2022).

The impact of weather on freight truck crash distribution underscores the need for weather-responsive traffic management strategies and prevention infrastructure design that considers seasonal changes to improve safety. The interaction between traffic status and weather conditions further complicates this dynamic (Okafor *et al.*, 2022). Adverse weather conditions impact traffic flow, leading to congestion and reduced visibility, which increases crash risks for freight trucks. Therefore, understanding the interplay between these factors is essential for developing comprehensive traffic management strategies that enhance road safety.

Socioeconomic and demographic factors are often correlated with transportation infrastructure quality and traffic conditions, therefore impacting the spatial distribution of freight truck-involved crashes (Yuan *et al.*, 2023; Yuan & Wang, 2021). Areas with high economic activity and dense industrial and logistical land use typically experience more freight traffic, leading to increased crash risks (Yu *et al.*, 2024). Dumbaugh and Rae (2009) explored how urban design and land use patterns impact crash distribution, emphasizing the need for integrated planning that considers both economic and traffic dynamics. Demographic disparities have a critical role in shaping spatial distributions. Regions with higher concentrations of minority and low-income populations often face poorer road conditions and higher crash risks. Yu *et al.* (2024) have highlighted the disparities in road network patterns across different socioeconomic groups, suggesting targeted interventions to address these inequalities. These findings underscore the importance of equitable infrastructure development to enhance safety for all road users, particularly in underserved communities. The interplay between the factors underscores the complexity of freight truck crash dynamics and the need for novel approaches to consider this latent interplay.

*2.2 Freight truck-involved crashes estimation models*

Existing research on freight truck-involved crashes has employed various models to estimate and analyze crash occurrences, each capturing the complexity and dynamics of influencing factors, mainly including three primary groups: traditional statistical models, machine learning-based models, and hybrid models (Hossain *et al.*, 2024a). Statistical models have been the cornerstone of crash estimation for decades since the ability to identify and quantify relationships between crash occurrences and influencing factors through statistical methods (Yuan & Wang, 2021). Existing studies have adopted statistical models, such as the zero-inflated and negative binomial models, to overcome the sparse spatial distribution of freight truck-involved crash features (Lord *et al.*, 2005; Pew *et al.*, 2020). Yang *et al.* (2021a) have used the zero-inflated Poisson model to analyze the negative correlation relationship between the distance to freight transport centers and the fatal and evident injury crash density. The freight truck-involved crashes also presents spatial autocorrelation patterns, hence, Yuan and Wang (2021) have used the spatial lag model (SLM) to model the spatial pattern between different demographic censuses and provided solid evidence to the spatial equity issues of freight truck-involved crashes.

While statistical models provide clear and interpretable results, they often rely on assumptions of linearity and independence among factors, which may not fully capture the complex interactions inherent and the nonlinear relationship in crash data. Yu *et al.* (2024) proposed a representative norm in this regard. They have utilized the Tweeide Boost machine learning models and combined them with the freight facilities extracted from the high-resolution satellite imagery to overcome the difficulties of addressing the zero-inflated issue in machine learning (Yu *et al.*, 2024). A previous work by them offered another approach, which provided a concept of freight truck crash hazards (FTCHs) to express the latent hazards of the crashes and further designed Bayesian deep learning models to understand road network patterns' role in shaping local truck crashes' severity (Yuan *et al.*, 2023). Different from the above-supervised machine learning methods, combined with the interpretable framework, such as Shap theory (Yang *et al.*, 2021a), Hossain *et al.* (2024b) employed an unsupervised learning algorithm to analyze crash data, revealing distinct patterns and spatial clusters that identify high-risk crash areas at high-speed locations. Furthermore, recent studies have attempted to combine the advantages of traditional statistical models with advances in machine learning-based methods. Hossain *et al.* (2024a) employ a two-step hybrid modeling approach using Extreme Gradient Boosting to identify key variables and Correlated Random Parameter Ordered Probit with Heterogeneity in Means (CRPOP-HM) to predict crash injury severity, concluding that factors such as collision type, contributing factors, presence of passengers, location, and weekend driving significantly increase the risk of severe crashes involving older drivers on high-speed roads.

While these models provide significant insights into freight truck-involved crash estimation by capturing complex patterns and identifying key risk factors, many of the current methodologies, particularly in machine learning, face challenges when it comes to counterfactual inference. The ability to simulate "what-if" scenarios and understand causality is crucial for identifying the causes of freight truck-involved crashes and designing effective interventions. Meanwhile, although machine learning models excel in handling large datasets and capturing nonlinear relationships, they typically operate as "black boxes," lacking the interpretability required for causal inference. This gap in understanding the causal mechanisms behind crash occurrences limits the ability to predict the effects of potential interventions accurately. Hence, our research aims to address these challenges by integrating counterfactual inference techniques within a deep learning framework. By leveraging advanced modeling approaches, we seek to improve the estimation of freight truck crash risks and uncover the causal relationships driving these incidents. This will provide more actionable insights for policymakers and urban planners, ultimately contributing to enhanced safety measures and more effective strategies to reduce freight truck-involved crashes.

# 3 Methodology

## 3.1 Deep counterfactual inference model

To address the challenges of understanding and quantifying the effects of various factors on freight truck crashes, we have proposed deep counterfactual inference (DCI) models, designed to simulate hypothetical scenarios (counterfactuals) to estimate how changes in specific factors might influence crash occurrences for prevention. The core of the DCI model is to predict the difference in the probability distribution between the actual severity of freight truck-related crashes under any combination of influencing factor changes at the time of the accident as counterfactual scenarios. By simulating counterfactual scenarios and estimating the differences with the factual outcome, the model allows us to estimate how changes in conditions or interventions might influence crash occurrences. Specifically, we introduce the DCI's network structure first and then explain the mathematical modeling process of the DCI model's loss function and the implementation framework.

### 3.1.1 DCI's network structure

DCI model is proposed based on the multi-task deep learning framework, which consists of four components: the input component, the calibration component, the shared representation component, and two task-specific heads for the factual scenario and the counterfactual scenario respectively, as shown in Fig.1.

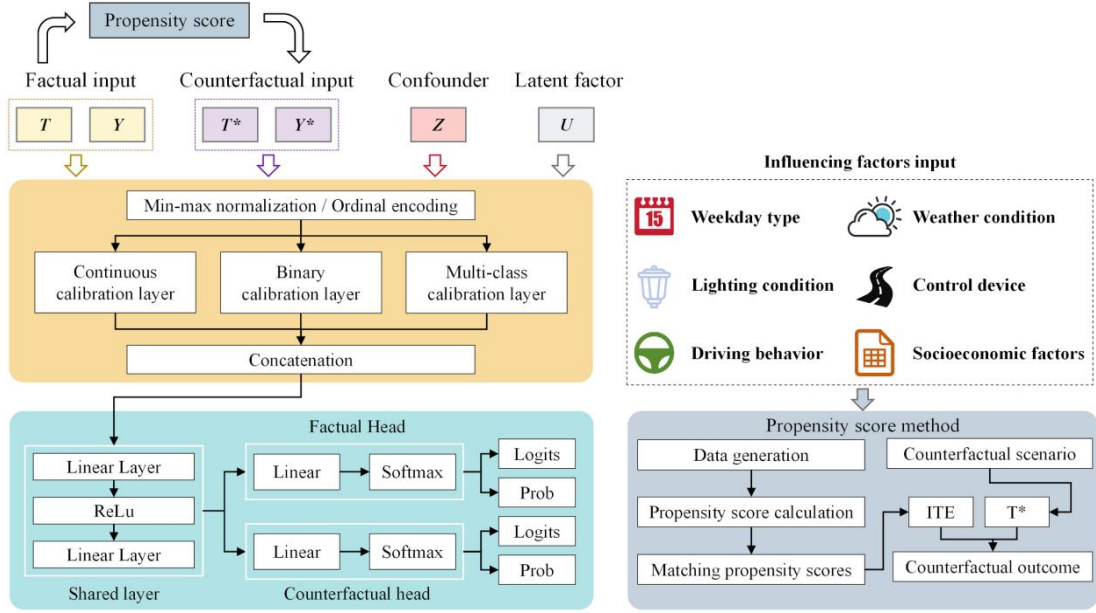

**Fig. 1** The architecture of the deep counterfactual inference model.

For each freight truck-related crash occurrence $i$, we have an outcome variable $\mathbf{Y}_i$ denoting the severity of the crash in an ordered form and covariates $\mathbf{X}_i$, a vector representing the observed conditions and influencing factors at the time of the crash (e.g., Lighting condition, weather condition, control device condition). There are two categories in the covariates $\mathbf{Z}_i$, one is the confounders $\mathbf{X}_i$, influencing not only the outcome but also other covariates, the other category is treatment variables $\mathbf{T}_i$. And we have preliminary counterfactual inference results $\mathbf{Y}_i^*$ and $\mathbf{T}_i^*$ for the collaboration learning process of the counterfactual scenario in the DCI model, obtained from methods like propensity score matching, providing initial estimates of counterfactual outcomes and treatments. In addition, we use $\mathbf{U}_i$ to denote unobserved factors influencing freight truck crashes.

The input to the model includes observed variables and preliminary counterfactual inference results, denoted as $\left[ \mathbf{T}_i, \mathbf{T}_i^*, \mathbf{X}_i, \mathbf{U}_i \right], \forall i \in I$. Considering the complexity of factors causing freight truck-related crashes (Shinar & Hauer, 2024; Yu *et al.*, 2024), which include binary variables and ordinal variables that directly represent immediate causes, as well as environmental factors represented by continuous variables (Haddad *et al.*, 2023; Hong *et al.*, 2019), we employ a calibration component in the model. This component helps to map continuous variables, binary variables, and ordinal variables to a standardized range to ensure consistency and facilitate learning, as presented in Fig. 1. Specifically, the continuous variables are scaled to a standardized range (e.g., [0, 1]) using min-max normalization, see Eq. (1). And for an ordinal variable with *K* categories ranked from 1 to *K*, the component works as Eq. (2). The calibration ensures that all input features are on a comparable scale, facilitating the training process and improving model performance.

$$\hat{x}_i = \frac{x_i - \min(x)}{\max(x) - \min(x)}, \forall x_i \in \mathbf{X}_i \cup \mathbf{T}_i \tag{1}$$

$$\hat{x}_i = \frac{x_i - 1}{K - 1}, \forall x_i \in \mathbf{X}_i \cup \mathbf{T}_i \tag{2}$$

To capture the nonlinear and high-dimensional relationships among these freight truck-related crash causes, we introduce the shared representation components, which consist of a series of fully connected layers, allowing the model to learn a unified, high-level representation of the calibrated inputs, and enabling the model to integrate various influencing factors and extract meaningful patterns essential for accurate predictions. Since the interaction and nonlinear relationship between causes have homogeneity in the factual and counterfactual scenarios, the fully connected layers are designed into two parts, i.e., the shared representation components and the task-specific modules. The latent representation of the shared representation components can be donated as Eq. (3).

$$H_i = \phi\left(W_2 \cdot \phi\left(W_1 \cdot \widehat{X}_i + b_1\right) + b_2\right) \tag{3}$$

where $\widehat{X}_i$ denotes the concatenated calibrated input vector; $W_1, W_2$ are weight matrices and $b_1, b_2$ represent bias vectors; $\phi(\cdot)$ is the activation function. After obtaining the shared representation $H_i$, the model branches into task-specific modules. The factual outcome head predicts the probability distribution over $K$ severity levels under the observed conditions while the counterfactual outcome head predicts the probability distribution under alternative scenarios, as Eq. (4) and Eq. (5) respectively.

$$\widehat{\mathbf{Y}}_i^F = \mathrm{softmax}\left(f_{Factual}\left(H_i\right)\right) \in R^K \tag{4}$$

$$\widehat{\mathbf{Y}}_i^{CF} = \mathrm{softmax}\left(f_{Counterfactual}\left(H_i\right)\right) \in R^K \tag{5}$$

where $f_{Factual}$, $f_{Counterfactual}$ are neural network functions, similar as Eq. (3), (e.g., fully connected layers) that output logits for each of the $K$ severity levels. The SoftMax function ensures that the outputs are valid probability distributions over the $K$ classes.

### 3.1.2 Loss function

To train the proposed model, we define a composite loss function that combines the discrepancies between the predicted and actual outcomes, as well as regularization terms to account for latent variables, as Eq. (6).

$$Loss_{DCI} = Loss_{Factual} + \gamma \cdot Loss_{Counterfactual} + \lambda \cdot Loss_{Reg} \tag{6}$$

where the loss function of the factual scenario of the freight truck-related crashes $Loss_{Factual}$ is written as Eq. (7) using categorical cross-entropy loss. $Loss_{Counterfactual}$ in Eq. (8) measures the error in predicting the preliminary counterfactual estimates $Y_i^*$ using categorical cross-entropy loss as well. And $Loss_{\text{Reg}}$ is designed to mitigate the influence of unobserved confounders $U_i$. Hyperparameters $\lambda, \gamma$ are employed to control the importance of the counterfactual loss and regularization term, respectively. The implementation framework outlining the steps of the model is as in Appendix A.

$$Loss_{\text{Factual}} = -\frac{1}{N}\sum_{i=1}^{N}\sum_{k=1}^{K} y_{ik} \cdot \log\left(\widehat{y_{ik}^{F}}\right) \tag{7}$$

$$Loss_{\text{Counterfactual}} = -\frac{1}{N}\sum_{i=1}^{N}\sum_{k=1}^{K} y_{ik}^{*} \cdot \log\left(\widehat{y_{ik}^{CF}}\right) \tag{8}$$

### 3.1.3 Estimating counterfactual effects

After training, our DCI model enables us to estimate the impact of changing any combination of factors $X \in T$ on the probabilities of various crash severity levels. Since the outputs are probability distributions over severity levels, we can address two key questions to denote the counterfactual effects: (1) Does changing a given set of factors $X \in T$ cause the freight truck-related crash's severity level to change between factual and counterfactual scenarios? (2) If the severity level remains the same between factual and counterfactual scenarios, does the change in factors $X \in T$ alter the probabilities associated with that severity level?

For the first question, we define the severity level of individual treatment effect (ITE) $\widehat{ITE}_{i,level}\left(X\right)$, which indicates whether the predicted severity level changes when the influencing factors $\mathbf{X}$ are altered. We determine the severity level by selecting the one with the highest predicted probability in both the factual and counterfactual scenarios. Mathematically, this can be expressed as Eq. (9). The corresponding average treatment effects (ATE) can be denoted as Eq. (10).

$$\widehat{ITE}_{i,level}\left(X\right) = \arg\max_{k\in K}\widehat{y_{i,k}^{CF}} - \arg\max_{k\in K}\widehat{y_{i,k}^{F}} \tag{9}$$

$$\widehat{ATE}_{level} = \frac{1}{N}\sum_{i=1}^{N}\widehat{ITE}_{i,level}\left(\mathbf{X}\right) \tag{10}$$

where $\widehat{y_{i,k}^{CF}}, \widehat{y_{i,k}^{F}}$ are the predicted probabilities for severity level $k$ of the factual and counterfactual scenarios, respectively. $N$ denotes the sample number.

For another question, when the severity level does not change ($\widehat{ITE}_{i,level}\left(\mathbf{X}\right) = 0$), we introduce the probability difference individual treatment effect $\widehat{ITE}_{i,probability}\left(\mathbf{X}\right)$ to represent the counterfactual effect. This measures the difference in the predicted probabilities of the current same severity level between the factual and counterfactual scenarios, which can be defined as Eq. (11). The corresponding ATE can be calculated as Eq. (12).

$$\widehat{ITE}_{i,probability}\left(\mathbf{X}\right) = \max_{k\in K}\widehat{y_{i,k}^{CF}} - \max_{k\in K}\widehat{y_{i,k}^{F}} \tag{11}$$

$$\widehat{ATE}_{probability} = \frac{1}{N}\sum_{i=1}^{N}\widehat{ITE}_{i,probability}\left(\mathbf{X}\right) \tag{12}$$

By utilizing these two counterfactual indexes, we can comprehensively assess the counterfactual effects of changing influencing factors on both the severity level and the associated probabilities of freight truck-related crashes.

### 3.2 Outcome, treatment, and confounder variables

This analysis of freight truck-involved crashes organizes variables into three key categories: outcome, treatment, and confounder variables (Table A3). These categories form a causal framework that helps to understand the factors influencing crash severity, with a focus on potential spatial justice concerns and disparities in crash outcomes.

The crash severity level is chosen as the outcome variable, which measures the injuries degree resulting from freight truck crashes, ranging from minor injuries to fatalities (Taylor *et al.*, 2018). The crash severity level contributes to understanding the freight truck-related accident consequences and assessing the effectiveness of interventions (Shin, 2024a). By identifying and quantifying the factors that cause higher crash severity, policymakers can develop more targeted and effective safety measures, especially in regions with potential socio-economic or road infrastructural spatial justice concerns.

The treatment variables encompass factors that have a direct causal effect on crash severity level and could be used to make intervention policies, including road environment, driving behavior, and the involvement of road users (Das *et al.*, 2022; Hong *et al.*, 2019). Road environment factors, such as lighting conditions, control device conditions, and weather conditions, significantly influence visibility and safety on the road, particularly for freight trucks which require more distance to stop and are harder to maneuver (Uddin & Huynh, 2017). Poor lighting, malfunctioning traffic signals, and adverse weather conditions can exacerbate the risks associated with truck operations, leading to more severe crashes. Driving behavior also plays a critical role in crash severity levels. Factors such as improper turning involvement and alcohol/drug involvement can be mitigated through stricter law enforcement, public education campaigns, and improved traffic management systems, which should be considered in policy formulation (Newnam & Goode, 2015). Furthermore, the involvement of pedestrians, cyclists, and motorcyclists in freight truck crashes presents a heightened risk due to the vulnerability of road users (J. Wang *et al.*, 2022). Protecting them through dedicated infrastructure, such as safer crosswalks and designated lanes should inform policy decisions.

The confounder variables include socioeconomic, land use, and road network attributes which impact crash severity levels indirectly (Appendix B Table A2-A3). These variables are crucial for identifying spatial justice concerns, as they often reveal disparities in how different regions are impacted by freight truck crashes. Socio-economic characteristics such as population density, mean household income, and minority percentage provide a view into regional crash severity differences (Yuan & Wang, 2021). Higher population density may increase traffic interactions and raise the likelihood of severe accidents, while lower-income and minority-dense areas often face insufficient investment in road infrastructure, making them more susceptible to severe crashes.

Land use attributes, including the proportions of service sector, industrial, retail trade, and transportation/warehousing jobs, reflect the intensity of freight-related activity in specific regions (Shin, 2024b). Areas with high concentrations of freight traffic, particularly those linked to transportation and warehousing, are at a higher risk of severe crashes due to the increased presence of heavy vehicles on the roads. Lastly, road network structure variables, such as average road segment length and intersection density, capture key elements of road design that influence crash risk (Kang, 2023). Higher intersection density and shorter road segments tend to increase the likelihood of vehicle interactions, raising the potential risk for severe crashes.

# 4. Study area, data, and modelling procedure

## 4.1 Study area and data sources

This research employs the Los Angeles Greater Metropolitan Area (LAMA), which spans over 33,000 square miles and includes a population of approximately 18.7 million people (Yoon *et al.*, 2018) as the study area and uses ten years of freight truck-related crash data from 2012 to 2022, obtained from the California Statewide Integrated Traffic Records System (SWITRS) database (the Southern California Association of Governments, 2022). The total number of freight truck-related crashes is 28,626. Table 1 shows the proportions of fatalities and injuries in these crashes. Among them, there were a total of 1,015 fatal crashes, resulting in 1,231 deaths, with the most severe crash causing 6 deaths. Of the 28,626 crashes, 27.7% resulted in at least two or more injuries, with the most severe crash causing 34 injuries. SWITRS classifies crash severity into four levels: fatal injury, suspected serious injury or severe injury, suspected minor injury or visible injury, and possible injury or complaint of pain. In our study, fatal injury crashes numbered 1,015, accounting for 3.55% of all crashes; suspected serious injury or severe injury crashes totaled 2,179, making up 7.61%; suspected minor injury or visible injury crashes were 8,526, representing 29.80%; and possible injury or complaint of pain crashes amounted to 16,904, comprising 59.04% of the total.

**Table 1.** Distribution of freight truck-related crashes by fatalities and injuries number.

| Number of Deaths/ Injuries per Crash | Fatal crash | | Injury crash | |
|---|---|---|---|---|
| | Number of Crashes | Proportion (%) | Number of Crashes | Proportion (%) |
| 0 | 27,611 | 96.45% | 731 | 2.55% |
| 1 | 940 | 3.28% | 20,689 | 72.29% |
| 2 | 62 | 0.22% | 5,005 | 17.49% |
| 3 | 8 | 0.03% | 1,478 | 5.16% |
| 4 | 3 | 0.01% | 472 | 1.65% |
| 5 | 1 | 0.00% | 179 | 0.63% |
| 6+ | 1 | 0.00% | 72 | 0.25% |
| Total | 28,626 | 100% | 28,626 | 100% |

We conducted a hotspot analysis of the spatial locations where crashes occurred, as shown in Fig. 2. Fig.2 presents that the spatial pattern of freight truck-related crashes exhibits significant inequity issues, with notable clustering phenomena appearing in downtown areas and along highways.

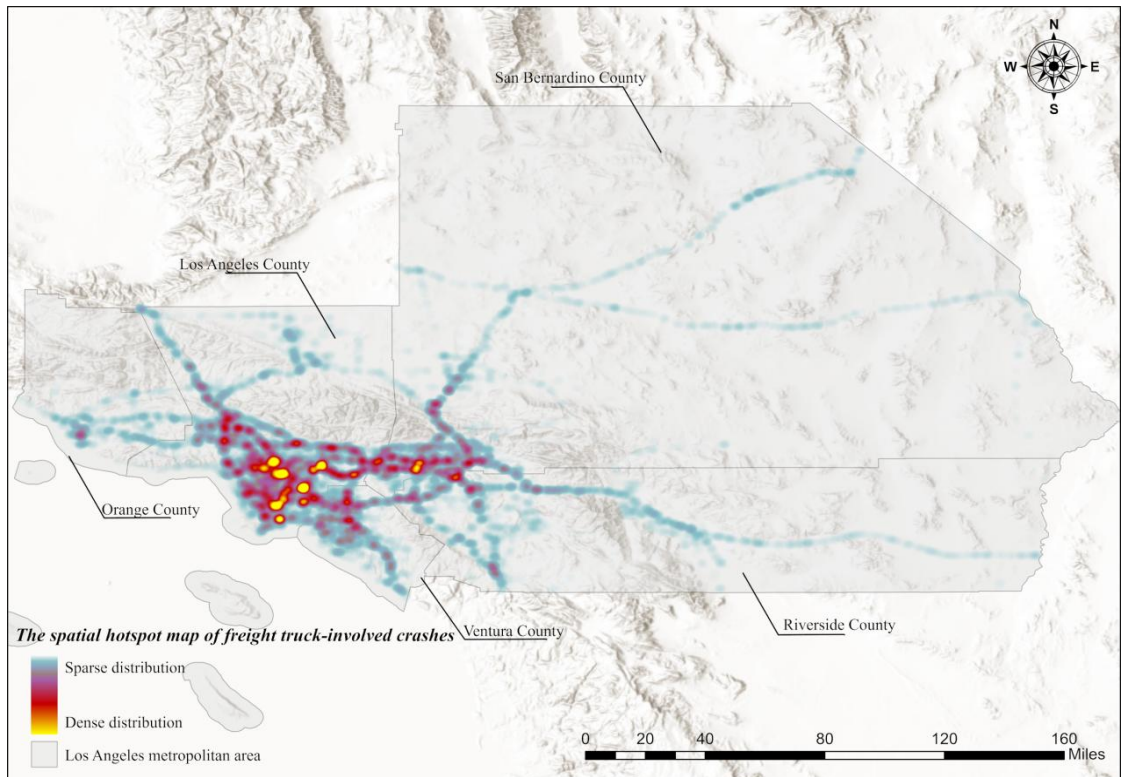

**Fig. 2.** The spatial hotspot map of the freight truck-involved crashes in the Los Angeles metropolitan area.

For the independent variables (Table A2), the direct cause variables of each freight truck-related crash are from the records in the SWITRS database. These variables include driver behavior-related factors (Improper turning involvement, alcohol or drug involvement), environmental and road conditions (e.g., measures of lighting condition and weather condition), road facilities and control (control device condition), and temporal and regional information (weekday type). In addition, we incorporated the spatial information of socioeconomic features and road networks as indirect causes to investigate the spatial injustice of freight truck-related crashes. The socioeconomic features used in this study were derived from the American

Community Survey (ACS) datasets provided by the U.S. Census Bureau for 2012 to 2022. For each crash occurrence, we utilized the ACS data corresponding to the specific year the crash took place, ensuring temporal alignment between the socioeconomic data and the crash data. The road network pattern indicators were obtained using the OSMnx toolkit, a Python package for retrieving, modeling, analyzing, and visualizing street networks from OpenStreetMap (OSM) data. By employing OSMnx, we extracted detailed road network data from OpenStreetMap and calculated various network indicators. The explanation of variable coding and descriptive statistics is shown in Appendix B. These variables constitute our set of independent variables used to analyze the factors influencing each accident occurrence.

*4.2 Model configuration and parameter settings*

The parameter configuration of the deep counterfactual inference model is shown in Table 2. The model's architecture is structured with shared layers, followed by two distinct branches—one handling factual predictions and the other dedicated to counterfactual outcomes. Both branches are configured identically, each comprising 256 hidden units to facilitate robust feature processing. This setup ensures that the model can effectively capture the complex patterns present in the crash data, which are often shaped by a combination of weather conditions, driver behavior, and road characteristics. The shared layers, composed of several dense layers with 128 neurons per layer, play a critical role in learning the intricate relationships between these input factors. The use of the ReLU activation function further enhances the model's ability to handle non-linearity, while simultaneously addressing the gradient vanishing problem. To address the issue of overfitting, a dropout rate of 0.3 is implemented, where a portion of neurons are randomly deactivated during each training iteration.

This process forces the network to generalize better, ensuring that it does not become overly reliant on any specific set of features. Furthermore, the model employs He initialization, which maintains the stability of gradients across layers, thus ensuring more consistent learning.

The model's loss function incorporates both factual and counterfactual loss components. The factual prediction loss is given greater importance with a weight of $\lambda_1 = 0.65$, emphasizing the need for accurate predictions in observed data. In contrast, the counterfactual ordering loss which captures the rank order of outcomes in hypothetical scenarios, is assigned a weight of $\lambda = 0.35$. This balance ensures that while prediction accuracy remains the priority, the model also effectively captures the nuances of alternative scenarios. Moreover, the regularization with a coefficient of $\lambda_{reg} = 0.01$ is applied to manage model complexity, reducing the risk of overfitting by penalizing large weights. We use the Adam optimizer, which dynamically adjusts the learning rate based on the first and second moments of the gradients. With an initial learning rate $\alpha = 0.001$, the model begins with relatively large updates, which gradually decrease as the training progresses, enabling fine-tuning. The batch size is set at 64, and the model is trained over 200 epochs, allowing sufficient time for convergence. An early stopping mechanism is employed to mitigate the risk of overfitting, ensuring that training halts when validation performance stabilizes.

The model accounts for unobserved confounding factors by incorporating latent variables $U$. These variables are designed to capture the influence of factors that are not directly observed but still impact crash severity outcomes. The latent variables are modeled with a dimension of $U_d = 10$ and are assumed to follow a normal distribution $U \sim N(0,1)$. This allows the model to incorporate hidden causal factors, thereby improving the robustness and accuracy of its causal inferences, both for factual and counterfactual scenarios.

**Table 2.** Parameter configuration of the deep counterfactual inference model.

| Parameter Name | Description | Value |
|---|---|---|
| Number of Dense Layers (L) | Number of fully connected layers in the shared representation component of the network. | 4 |
| Neurons numbers (N) | Number of neurons in each dense layer, representing model capacity to capture data patterns. | 128 |
| Dropout rate (D) | Probability of deactivating neurons during training to reduce overfitting. | 0.3 |
| Hidden units(H) | Number of hidden units in factual and counterfactual branches for processing predictions. | 256 |
| Output neurons (O) | Neurons correspond to the four crash severity levels (fatal, severe, minor, and possible injuries). | 4 |
| Regularization (L$_2$) | L$_2$ regularization applied to task-specific layers to penalize large weight magnitudes. | 0.01 |
| Learning rate (α) | Initial step size for gradient updates, decayed over time for stable convergence. | 0.001 |
| Batch size (B) | Number of samples processed before model weight updates. | 64 |
| Epochs (E) | Number of complete passes through the training dataset. | 200 |
| Factual loss weight (λ$_1$) | Weight balancing the contribution of the factual scenario to the overall loss function (prediction loss). | 0.65 |
| Counterfactual ordering loss weight (λ$_2$) | Weight emphasizing the importance of counterfactual ordering loss during training. | 0.35 |
| Regularization weight (λ$_{reg}$) | Regularization weight controlling the penalty applied to model weights. | 0.01 |
| Latent variable dimension (U$_d$) | Latent variable dimension representing unobserved factors influencing crashes. | 10 |

# 5 Results

## 5.1 Spatial distribution of fatal and injury-related freight truck crashes in the Los Angeles metropolitan region

The fatal and injury-related freight truck crashes in the Los Angeles metropolitan region are illustrated in Fig. 3, which reveals the substantial disparities in the geographic distribution and underscores concerns about spatial justice. Central urban areas and communities along major freight transportation corridors are the primary regions bearing more freight truck crashes. Specifically, areas characterized by high population density such as central Los Angeles County and its adjacent communities are significantly more affected by more crashes, accounting for 53.4% of freight truck crashes from 2012 to 2022. Conversely, the maps indicate that rural and sparsely populated regions, including parts of Ventura, San Bernardino, and Riverside Counties, exhibit lower crash densities. However, crashes in these regions have higher crash severity levels, with 1.243% fatal injuries and 1.441% severe injuries more than those in Los Angeles, probably due to the limited access to emergency services.

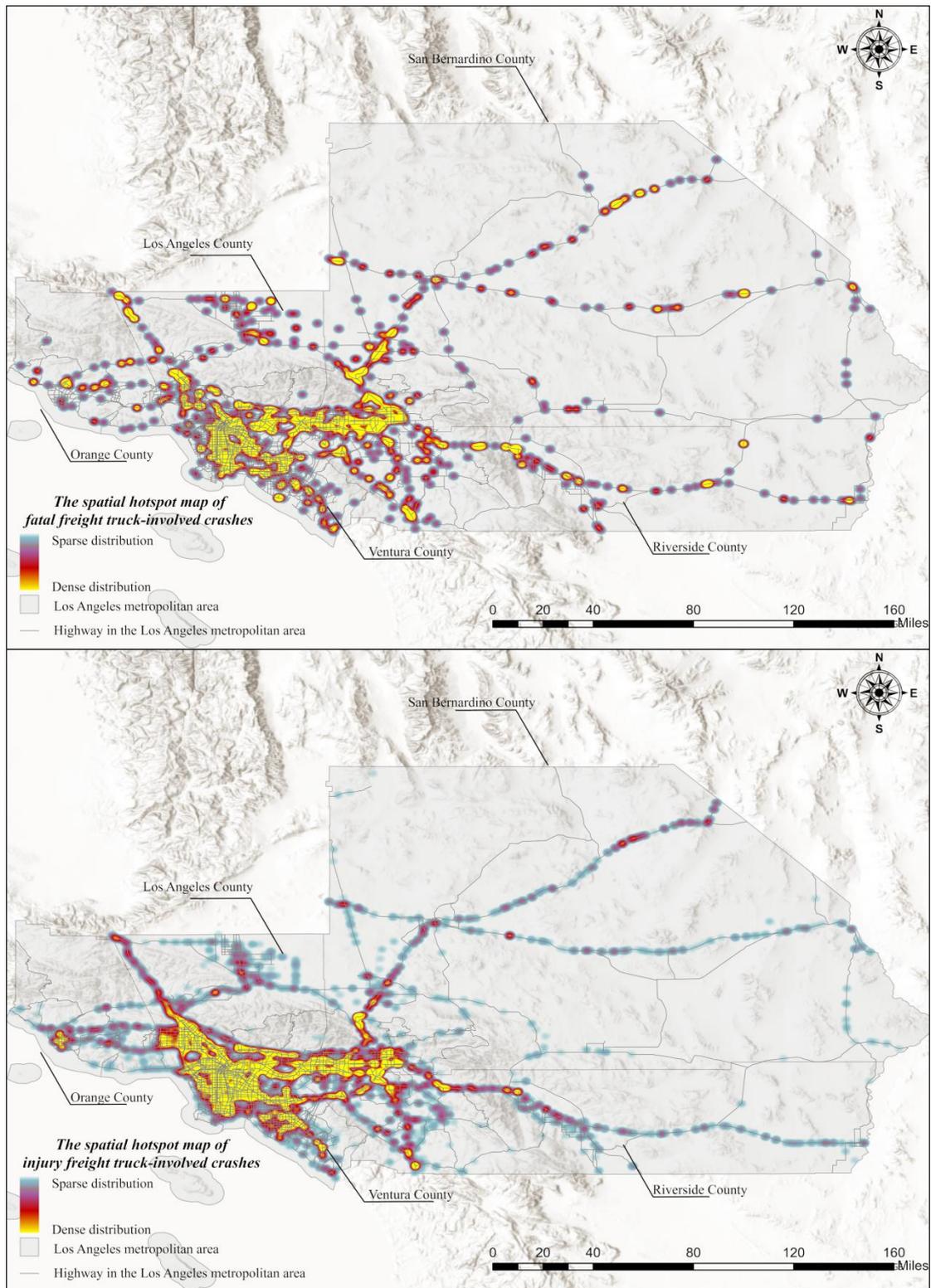

**Fig. 3.** The spatial hotspot map of the **(a.)** Fatal freight truck-involved crashes; **(b.)** Crashes caused at least 2 people injury; in the Los Angeles metropolitan area.

The distribution of freight truck crashes by severity levels across different mean household income groups also raises significant spatial justice concerns. In low-income areas, 537 accidents resulted in fatal injuries, and 1,152 accidents led to severe injuries. In comparison, high-income areas reported 478 fatal injury crashes and 1,027 severe injury crashes in total. These figures indicate that lower-income communities experience a higher number of both fatal and severe injuries, reflecting potential disparities in road infrastructure, emergency services, and traffic management.

When examining the distribution of crash severity levels across land-use characteristics, similar spatial justice concerns emerge, as demonstrated in Table 3. For regions with high retail trade job proportion, fatal injuries account for 3.56%, slightly higher than the 3.49% observed in low-retail job proportion areas. Severe injuries show an even more pronounced difference, with 7.77% in high-retail job proportion regions compared to 6.88% in low-retail job proportion regions. This disparity suggests that areas with dense retail activity may face greater risks of both fatal and severe injuries, likely due to increased pedestrian traffic or frequent interactions between pedestrians and freight vehicles. Moreover, in regions with a high transportation and warehousing job proportion, the fatal injuries rose to 3.78%, notably higher than the 3.25% observed in areas with lower transportation job concentrations. The statistics demonstrate that exposure to freight-related traffic in these regions underscores significant spatial justice concerns, as communities with more transportation and warehousing sectors disproportionately bear the burden of severe crashes.

**Table 3.** Crash severity levels by land-use characteristics.

| Land-use characteristics | Group | Fatal injury | Severe injury | Minor injury | Possible injury |
|---|---|---|---|---|---|
| Service sector job density | Low group | 3.38% | 7.66% | 29.80% | 59.16% |
| | High group | 4.24% | 7.41% | 29.73% | 58.62% |
| Industrial sector job density | Low group | 3.59% | 7.57% | 29.85% | 58.99% |
| | High group | 3.34% | 7.81% | 29.48% | 59.37% |
| Retail trade job density | Low group | 3.56% | 7.77% | 29.78% | 58.89% |
| | High group | 3.49% | 6.88% | 29.80% | 59.84% |
| Transportation/warehousing job density | Low group | 3.25% | 7.63% | 29.74% | 58.85% |
| | High group | 3.78% | 7.59% | 29.85% | 59.31% |

*5.2 Prediction performance in factual and counterfactual scenarios*

To assess the prediction performance of the proposed deep counterfactual inference model, a comparison was made with three alternative methodologies: Causal Forest (CF), Double Machine Learning (DML), and Generative Adversarial Networks (GANs). Causal Forest employs decision trees to estimate individual treatment effects through recursive partitioning (Wager & Athey, 2018). It divides the covariate space into subgroups, where the relationship between covariates and treatment effects can be considered homogeneous. For each partition, the average treatment effect is calculated based on leaf-level observations. The model captures the conditional average treatment effect by aggregating treatment effect estimates across multiple trees, enabling non-parametric and flexible handling of heterogeneous causal effects.

Double Machine Learning for causal inference focuses on the estimation of low-dimensional causal parameters while accounting for the high-dimensional nuisance parameters (Chernozhukov *et al.*, 2018). By applying regularization methods, DML addresses the complexity introduced by these nuisance parameters. To reduce the estimation bias, the DML further incorporates Neyman-orthogonal scores which ensure reduced sensitivity to errors in the nuisance parameters.

Generative Adversarial Networks (GANs) for counterfactual inference are built upon a generator-discriminator framework(Yoon *et al.*, 2018). The generator is responsible for producing counterfactual outcomes under varying treatment conditions, while the discriminator evaluates whether the generated outcomes are plausible within the given data context. Instead of directly aiming to match real-world results, the generator's goal is to generate counterfactual outcomes that align with the expected distribution based on a pre-established baseline model. The process is framed as a minimax optimization problem, where the generator seeks to minimize the discrepancy between the generated counterfactual outcomes and the expected outcomes under the reference model, and the discriminator maximizes this discrepancy. This iterative dynamic allows the generator to refine the quality of its counterfactual predictions while maintaining consistency with the underlying treatment effects and data structure.

The performance metrics comparison for crash severity level prediction in both factual and counterfactual scenarios in Table 4 demonstrates the superiority of the deep counterfactual inference model. In factual conditions, our model achieved the lowest MSE of 0.882, outperforming CF (1.971), DML (1.340), and GANs (1.039). Similarly, in counterfactual scenarios, our method maintained its performance with an MSE of 0.978, compared to CF(2.105), DML(1.516), and GAN(1.111). The results demonstrate the deep counterfactual inference model's ability to provide accurate predictions.

**Table 4.** Performance metrics comparison for crash severity level prediction in factual and counterfactual scenarios.

| Model | Scenario | MSE | RMSE | MAE |
|-------|----------|-----|------|-----|
| Causal Forest | Factual | 1.971 | 1.404 | 0.941 |
| Double Machine Learning | Factual | 1.340 | 1.158 | 0.640 |
| Generative Adversarial Networks | Factual | 1.039 | 1.019 | 0.492 |
| Our Method | Factual | 0.882 | 0.939 | 0.419 |
| Causal Forest | Counterfactual | 2.105 | 1.451 | 0.995 |
| Double Machine Learning | Counterfactual | 1.516 | 1.231 | 0.722 |
| Generative Adversarial Networks | Counterfactual | 1.111 | 1.054 | 0.532 |
| Our Method | Counterfactual | 0.978 | 0.989 | 0.466 |

Additionally, the performance results revealed a consistent trend across all four methods: each model performed better in predicting factual accident severity levels compared to counterfactual ones. It suggests that learning from observed accident data is inherently more straightforward, as factual labels represent real-world outcomes, whereas counterfactual labels involve the simulation of alternative scenarios. Despite the robustness of the deep counterfactual inference model in generating and analyzing counterfactual scenarios, it achieves higher accuracy in predicting factual outcomes.

*5.3 Individual causal effects of factors on freight truck crash severity*

For cases where crash severity levels change after treatment, the change value in severity is calculated as the Individual Treatment Effect (ITE). For accidents where the severity remains unchanged, the probability shift of remaining in this severity category is calculated as the ITE, as shown in Fig. 4. The Average Treatment Effect (ATE) for each factor is then derived by averaging the ITEs from both groups, as demonstrated in Table 5.

The involvement of pedestrians, bicyclists, and motorcyclists shows the highest ATEs. Pedestrian involvement has the largest impact, with an average severity level increase of 0.850 and an average probability increase of 45.312%, suggesting that freight truck accidents with pedestrians are more likely to result in severe outcomes. Similarly, Cyclist and Motorcyclist involvement (Average severity level increase of 0.449 and 0.714, respectively) demonstrate the greater risks for these road users. Besides, in regions with higher intersection density, the involvement of pedestrians, bicyclists, and motorcyclists led to an increase in the average severity level by 0.107, 0.193, and 0.092 respectively, compared to areas with lower intersection density, indicating the potential spatial justice problems where freight routes overlap with pedestrian and cycling paths.

The analysis of road environmental factors (Lighting condition, control device condition, and weather condition) demonstrates a notable impact on reducing freight truck crash severity levels. All three variables exhibit negative ATE values. Specifically, the lighting condition has an average severity level change of -0.267 and an average probability change of -14.386%, suggesting that enhanced road lighting improves visibility, thereby decreasing the crash severity level. Similarly, the control device condition reflected an average severity level change of -0.143 and an average probability change of -10.375%, demonstrating the importance of well-developed traffic control systems. Among road environmental factors, weather conditions exert the greatest influence, with an average severity level change of -0.325 and an average probability change of -16.742%, suggesting that favorable weather conditions are vital in mitigating accident severity. It is worth noting that low-income areas require more substantial improvements in road environmental factors compared to high-income areas. For instance, improving lighting conditions in low-income areas results in a 0.118 severity level decrease compared to high-income areas.

The analysis of driver behavior factors demonstrates a significant influence on increasing the crash severity. Improper turning involvement is associated with an average severity level change of 0.137 and an average probability change of 8.886%, indicating that improper turning contributes to the crash severity increase. Moreover, alcohol or drug involvement has a pronounced positive effect, with an average severity level change of 0.198 and an average probability change of 12.573%. The use of alcohol or drugs may lead to slower reaction times, poor judgment, and an increased likelihood of engaging in risky driving practices, all of which significantly exacerbate crash severity in freight truck accidents.

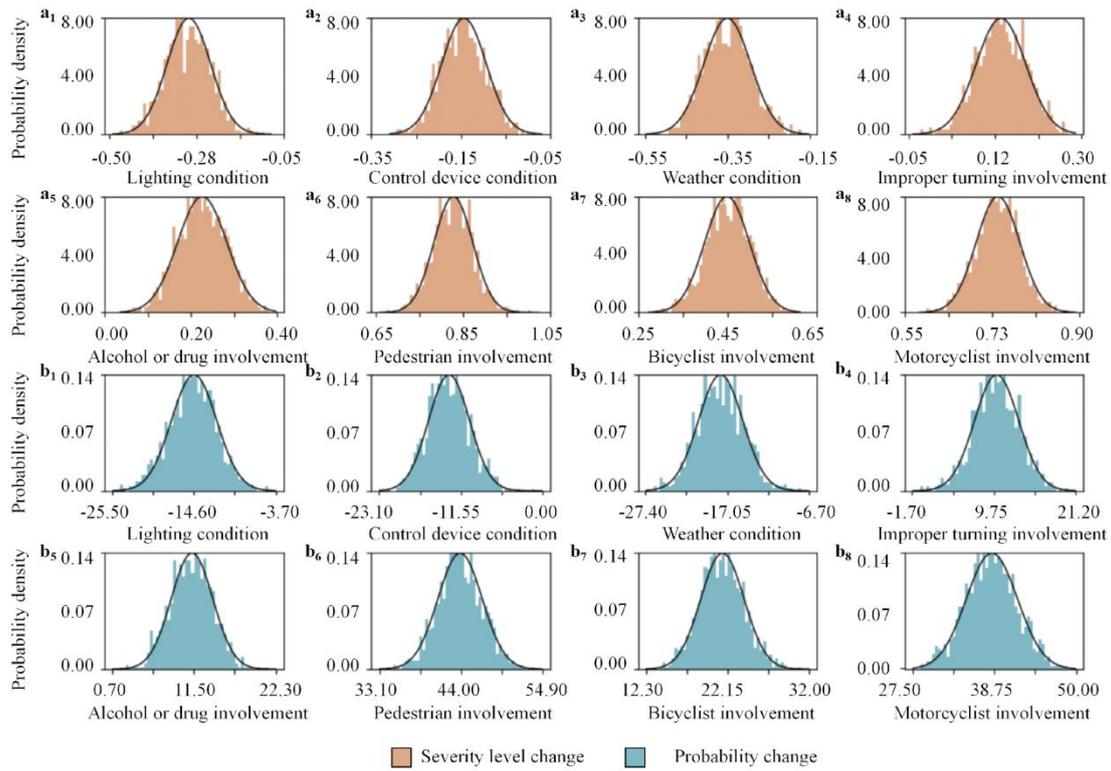

**Fig. 4.** Distribution of individual accident severity level and probability change. **(a)** Severity level change. **(b)** Probability change.

**Table 5.** Average treatment effect and for freight truck crash severity

| Treatment variable | Average severity level change | Average probability change |
|---|---|---|
| Lighting condition | -0.267 | -14.386 % |
| Control device condition | -0.143 | -10.375 % |
| Weather condition | -0.325 | -16.742 % |
| Improper turning involvement | 0.137 | 8.886 % |
| Alcohol or drug involvement | 0.198 | 12.573 % |
| Pedestrian involvement | 0.850 | 45.312 % |
| Bicyclist involvement | 0.449 | 22.887 % |
| Motorcyclist involvement | 0.714 | 38.532 % |

*5.4 Counterfactual inference on freight truck-involved crash*

Freight truck crashes often disproportionately affect certain regions due to socio-economic disparities and varying road conditions, raising potential spatial justice concerns. By leveraging counterfactual inference and control experiments, the impact of different intervention policies on crash severity is assessed across varying scenarios. The counterfactual inference results of lighting conditions across different population density groups are shown in Fig. 5. Regions with high population density (above 4,000 people/km²) exhibit the lowest crash severity compared to areas with medium and low population densities. For instance, the crash severity under dark conditions without street lights is 0.747 in high-density areas, while it rises to 0.753 in medium-density areas and 0.783 in low-density areas. The counterfactual inference results also reveal that regions with different population densities experience notable variations in average freight truck crash severity, with low-density areas exhibiting significantly higher severity compared to high-density regions. With adequate night-time lighting, the crash severity in high, medium, and low-density areas decreases to 0.659, 0.645, and 0.613, respectively, representing a reduction of 0.088, 0.108, and 0.170 in these areas.

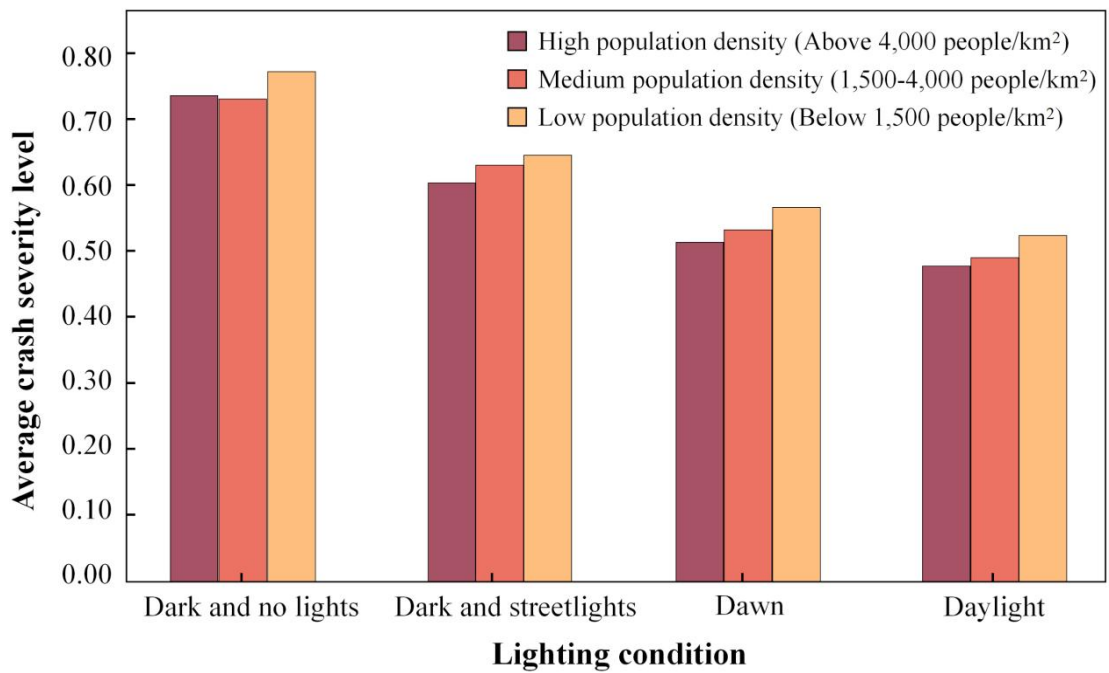

**Fig. 5.** Counterfactual inference result of lighting conditions among different population density groups.

The counterfactual analysis of weather conditions across different mean household income groups, as presented in Fig. 6, reveals a significant negative causal effect of weather on freight truck crash severity, with the impact being more pronounced in low-income areas. Under severe weather conditions, crash severity in low-income regions reaches 0.890, compared to 0.839 in high-income areas and 0.826 in medium-income areas. Even as weather conditions improve, the disparity in severity persists, with low-income regions experiencing higher severity levels of 0.856 in moderate weather and 0.563 in good weather, compared to 0.762 in moderate weather and 0.508 in good weather in high-income areas, respectively. These findings underscore that adverse weather conditions amplify crash severity more significantly in low-income regions, likely due to weaker road infrastructure and less effective traffic management systems.

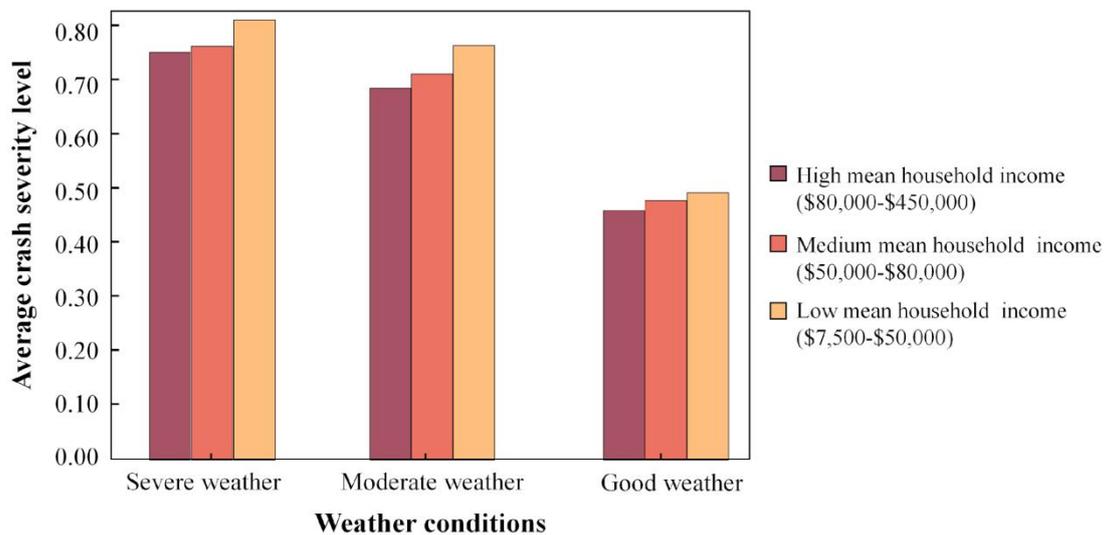

**Fig. 6.** Counterfactual result of weather conditions among different income groups.

The counterfactual results of the control device across varying minority percentage groups are presented in Fig. 7. Communities with a minority percentage exceeding 45.0% consistently show higher crash severity compared to areas with a lower minority percentage. In the absence of control devices, crash severity in high minority percentage regions reaches 0.703, while areas with a low or medium minority percentage report lower severity levels at 0.692 and 0.631, respectively. With functioning control devices, the disparity of freight truck-involved accident severity level persists with crash severity in areas with a high minority percentage at 0.535, compared to 0.528 in areas with a medium minority percentage and 0.516 in areas with a low minority percentage. However, in areas with a high minority percentage, functioning control devices lower crash severity by 0.168 compared to scenarios without devices, indicating that improving the functionality of control devices can significantly reduce this disparity.

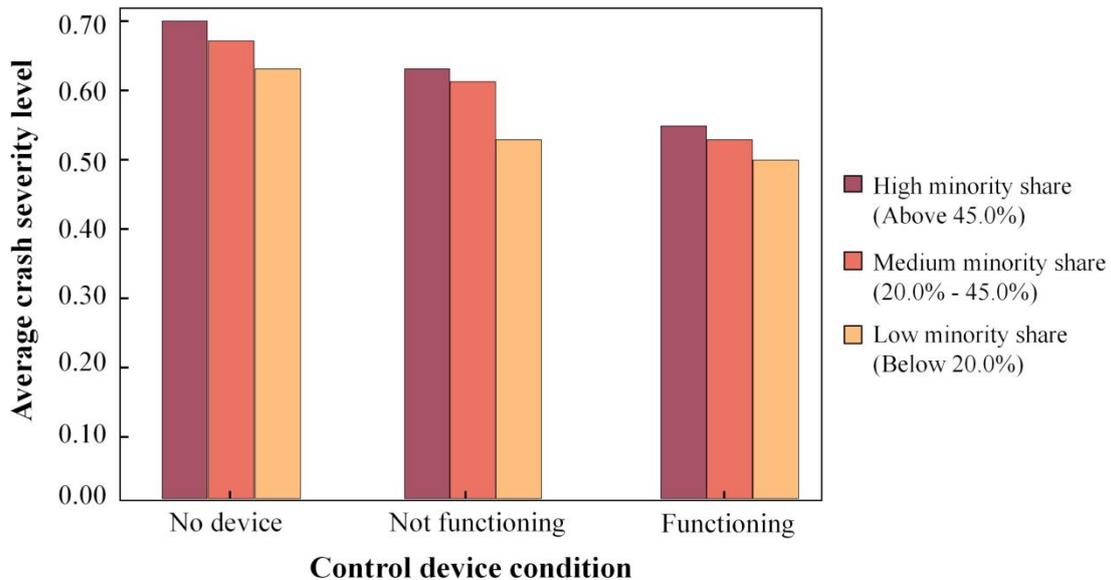

**Fig. 7.** Counterfactual inference result of control device among different minority shares groups.

The counterfactual inference result of pedestrian involvement across different levels of intersection density is presented in Fig. 8. Areas of high intersection density (40.0-160.0 intersections/km²) demonstrate a substantially higher crash severity of 1.574 when pedestrians are involved, compared to 1.349 in medium-density areas and 1.320 in low-density regions. Conversely, in the absence of pedestrian involvement, crash severity is significantly lower across all intersection density groups, with severity scores of 0.551 in high-density areas, 0.523 in medium-density areas, and 0.491 in low-density regions. This disparity reflects the heightened risk in densely intersected urban environments, where frequent interactions between freight trucks and pedestrians increase the likelihood of severe crashes.

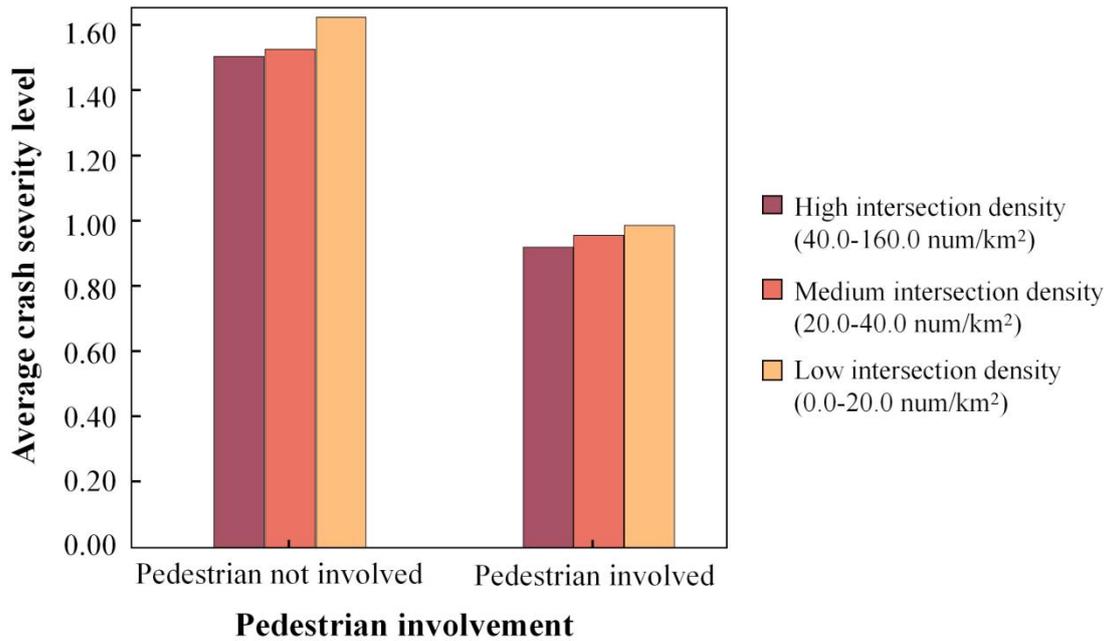

**Fig. 8.** Counterfactual inference result of pedestrian involvement among different intersection density groups.

## 6 Discussion

In response to the increasing demand for road freight transportation and the consequent rise in freight truck-related crashes, this study focuses on investigating the spatial justice concerns associated with these accidents. Through a deep counterfactual inference model, our analysis has shown that socio-economic and infrastructural factors contribute significantly to disparities in crash severity across the Los Angeles metropolitan area. Key findings indicate that low-population density, low-income, and minority-dense areas experience higher crash severity, often due to inadequate road infrastructure and environmental conditions. Our findings provide a detailed causal analysis of how the aforementioned factors affect freight truck-involved crashes, and highlight spatial inequities of minority group residential communities, underscoring the importance of interventions in enhancing both infrastructural and environmental attributes. Our contributions are twofold: First, by applying counterfactual inference, we move beyond traditional factor analysis to quantify the causal effects of road and environmental conditions on freight truck crash outcomes. Second, our findings highlight the spatial inequities in crash severity across regions, underscoring the demand for targeted safety interventions in vulnerable communities. This research not only deepens the understanding of the factors influencing freight truck crashes but also provides a crucial framework for policymakers to develop more equitable transportation safety strategies.

Our findings enrich the current scholarship in several key directions. We delve into the intricate dynamics of freight truck-related crashes and their spatial justice implications, drawing on established research to bolster our analytical framework. Our deep counterfactual inference model not only builds upon previous work that underscores the significance of socio-economic and infrastructural factors in determining crash severity (Uddin and Huynh, 2017), but also extends their findings by introducing a causal analysis that quantifies the impact of road and environmental conditions on crash outcomes. This approach allows us to move beyond traditional factor analysis, offering a more nuanced understanding of how specific conditions, such as inadequate lighting and weather-responsive technologies, contribute to the severity of crashes in low-population density areas. Furthermore, our research complements the work of McDonald et al. (2019) by advocating for resilient road infrastructure programs in low-income areas, emphasizing the importance of resurfacing roads and reinforcing embankments to enhance safety. We also align with Das et al. (2022) in recognizing the need for smart traffic signals in minority-dense communities to ensure the smooth flow of freight trucks, thereby addressing the spatial justice issues that persist despite the presence of traffic control devices. While our study shares the common goal with these esteemed researchers of improving road safety, our methodology provides a distinct advantage by integrating a comprehensive causal analysis that captures the interplay between various factors influencing crash severity. This deep counterfactual inference model allows us to quantify the causal effects, offering a more precise understanding of how targeted interventions can lead to more equitable transportation safety strategies. Our findings not only echo the concerns raised by these authors but also provide a detailed causal framework that can guide policymakers in developing targeted and effective interventions to reduce crash severity and promote spatial justice in road freight transportation.

Our study provides far-reaching policy implications to transport and urban planning to reduce the social inequality of safe risks in the less populated regions, low-income communities and minority concentrations. Firstly, in less populated areas where the freight truck crash severity is high, inadequate road infrastructure increases safety risks. These regions, often rural or suburban, suffer from a lack of essential safety features like lighting, traffic control devices, and proper road maintenance, making them more vulnerable to severe accidents. A key policy intervention is improving lighting along freight corridors and intersections, especially in underserved areas. Installing energy-efficient LED and solar-powered streetlights, equipped with adaptive technologies, can enhance visibility and reduce crash severity (Uddin and Huynh, 2017). Additionally, upgrading traffic control devices is critical. Implementing advanced traffic signal systems with real-time monitoring and adaptive controls can mitigate risks at key intersections with heavy freight traffic (Zhang et al., 2020). Traffic calming measures, such as speed bumps and roundabouts, can also slow vehicles and improve safety. Regular infrastructure audits and a comprehensive road safety program will ensure that these areas receive necessary maintenance and upgrades to prevent severe crashes.

Secondly, low-income communities are found to be disproportionately affected by adverse weather, leading to higher freight truck crash severity. These areas often suffer from poor drainage, degraded road surfaces, and a lack of weather-adaptive traffic management tools, making them more vulnerable. To address these issues, policy measures should prioritize resilient infrastructure programs focused on resurfacing roads, reinforcing embankments, and improving traction (McDonald et al., 2019). A broader Climate-Resilient Transportation Plan can target regions prone to extreme weather. Integrating weather-responsive technologies, such as real-time monitoring systems and dynamic speed regulations, can help reduce crash risks by adjusting speed limits based on conditions. Additionally, enhanced winter maintenance, including timely snow removal and de-icing, is critical for areas facing severe weather (Shin, 2024a). Freight route optimization systems should also divert trucks from hazardous routes, promoting long-term road safety and resilience in low-income regions.

Thirdly, the counterfactual results reveal a significant spatial justice issue in minority-dense communities, where freight truck crash severity remains elevated despite the presence of traffic control devices. These areas often suffer from underinvestment in infrastructure and road safety, requiring targeted interventions. Implementing comprehensive traffic management systems, such as smart traffic signals that adapt to real-time conditions, is essential for improving freight truck flow through residential areas (Das et al., 2022). Regular maintenance and upgrades of traffic control devices—signals, signage, and speed monitoring systems—are crucial for enhancing road safety. Automated enforcement systems, including speed and red-light cameras, can further reduce reckless driving.

Several limitations in our study should be considered to extend the current findings. First, although the deep counterfactual inference model effectively captures various road environmental factors and driver behaviors, the study did not incorporate real-time data. The static nature of the dataset used limits our ability to account for temporal fluctuations in traffic patterns that could influence crash severity. Future research could enhance this analysis by integrating real-time traffic data and temporal factors, providing a more accurate and dynamic understanding of how these conditions impact crash risks. This research primarily addresses spatial justice concerns through infrastructural and socio-economic lenses but does not fully incorporate the complexities of human factors, such as driver fatigue, distraction, or behavioral adaptation to infrastructure. While driver behavior-related variables, such as alcohol or drug involvement, were included, the model does not directly account for more nuanced driver-specific factors that could substantially influence crash outcomes. Future work could explore the integration of behavioral data or telematics-based monitoring to assess how individual driving patterns contribute to the overall severity level in different spatial conditions.

## 7 Conclusion

This study reveals significant spatial disparities in freight truck crash severity across the Los Angeles metropolitan area, rooted in socio-economic and infrastructural inequities. Using a deep counterfactual inference model, the research highlights critical concerns, such as inadequate lighting, a lack of real-time weather monitoring, and insufficient adaptive traffic management technologies in vulnerable communities. These disparities call for a shift from reactive measures to long-term, sustainable solutions. We advocate for that future strategies that integrate real-time data and dynamic modeling to adapt to changing traffic patterns and environmental conditions. Advanced technologies like adaptive lighting and dynamic speed regulation should be paired with regular upgrades of traffic control devices. Additionally, incorporating telematics-based monitoring could help better understand driver behavior and its impact on crash outcomes. The model and analytical farmwork implemented in our study are highly repeatable, reproduceable, and replicable to a broader context for transportation safety. By implementing data-driven and targeted interventions, our study provides possibility for policymakers to create a safer, more equitable road system, ensuring that all residents, regardless of socio-economic status, are prioritized in road safety efforts.

# Appendix A The implementation framework

**Table A1.** Pseudocode for deep counterfactual inference model (DCIM)

| |
|---|
| **Input:** Dataset D = (X$_i$, T$_i$, Y$_i$, T'$_i$, Y'$_i$, U$_i$) for i = 1 to N, hyperparameters $\lambda_1$, $\lambda_2$, $\lambda_{reg}$, $\alpha$ |

01. Initialize model parameters θ
02. Compute propensity scores P(T$_i$ | X$_i$) for each i using Propensity Score Matching (PSM)
03. For each epoch:
04.     For each batch B ⊆ D:
05.         For each sample i in batch B:
06.             // Step 1: Calibration
07.             Apply calibration to covariates X$_i$ and confounders Z$_i$ to standardize variables
08.             X̄$_i$ = Calibrate(X$_i$, Z$_i$)
09.
10.             // Step 2: Generate latent variable U
11.             U$_i$ = Sample U ∼ N(0, 1) or other distribution
12.
13.             // Step 3: Forward Pass
14.             Compute shared representation H$_i$ via shared layers
15.             H$_i$ = Shared_Layers([X̄$_i$, U$_i$])
16.
17.             Predict factual outcome Ŷ$_i^F$ using factual outcome head
18.             Ŷ$_i^F$ = Factual_Head(H$_i$, T$_i$)
19.
20.             Predict counterfactual outcome Ŷ$_i^{CF}$ using counterfactual outcome head
21.             Ŷ$_i^{CF}$ = Counterfactual_Head(H$_i$, T'$_i$)
22.
23.             // Step 4: Loss Computation
24.             Compute factual loss LF using cross-entropy loss with weight $\lambda_1$
25.             LF = $\lambda_1$ * CrossEntropy(Y$_i$, Ŷ$_i^F$)
26.
27.             Compute counterfactual loss LCF using cross-entropy loss with weight $\lambda_2$
28.             LCF = $\lambda_2$ * CrossEntropy(Y'$_i$, Ŷ$_i^{CF}$)
29.
30.             Compute regularization term LReg with weight $\lambda_{reg}$ to account for latent variable U
31.             LReg = $\lambda_{reg}$ * ||θ||²
32.
33.             Compute the total loss
34.             L = LF + LCF + LReg
35.
36.             // Step 5: Backward Pass
37.             Compute gradients ∇θL
38.
39.             // Step 6: Update Parameters
40.             Update model parameters θ using Adam optimizer
41.             θ = θ - $\alpha$ * ∇θL
42.
43. **Output:** Trained model capable of estimating Individual Treatment Effect (ITE) and Average Treatment Effect (ATE)
**Function Definitions:**
44. Function Calibrate(X, Z):
45.     // Standardizes covariates X and confounders Z using predefined calibration functions
46.     Return calibrated [X, Z]
47.
48. Function CrossEntropy(y_true, y_pred):
49.     // Computes cross-entropy loss between the true outcome and predicted outcome
50.     Return loss_value
51.
52. Function PropensityScoreMatching(X, T):

53.　　// Generates propensity scores for given covariates X and treatment T using logistic regression or machine learning models

54.　　Return propensity_scores

55.

56. Function AdamOptimizer(gradients, learning_rate):

57.　　// Updates model parameters using the Adam optimization algorithm

58.　　Return updated_parameters

## Appendix B Explanation of variable coding and descriptive statistics

**Table A2.** Discrete variables coding definitions.

| Variable | Coding definitions |
|---|---|
| Crash severity level | 3 = Fatal injury; 2 = Suspected serious injury or severe injury; 1 = Suspected minor injury or visible injury; 0 = Possible injury or complaint of pain; |
| Lighting condition | 3 = Dark and no lights; 2 = Dark and streetlights; 1 = Dawn; 0 = Daylight; |
| Control device condition | 2 = Functioning; 1 = Not Functioning; 0 = No device; |
| Weather condition | 2 = Good weather; 1 = Moderate weather; 0 = Severe weather; |
| Improper turning involvement | 1 = Improper turning involved; 0 = Improper turning not involved; |
| Alcohol or drug involvement | 1 = Alcohol or drug involved; 0 = Alcohol or drug not involved; |
| Pedestrian involvement | 1 = Pedestrian involved; 0 = Pedestrian not involved; |
| Cyclist involvement | 1 = Cyclist involved; 0 = Cyclist not involved; |
| Motorcyclist involvement | 1 = Motorcyclist involved; 0 = Motorcyclist not involved; |
| Weekday Type | Sunday = 6; Saturday = 5; Friday = 4; Thursday = 3; Wednesday = 2; Tuesday = 1; Monday = 0; |

**Table A3.** Descriptive statistics of the outcome variable and variables used as treatment and confounders.

| Variable | Type | Unit | Mean | Std | Min | Max |
|---|---|---|---|---|---|---|
| Crash severity level | Outcome | - | 0.56 | 0.78 | 0.00 | 3.00 |
| Lighting condition | Treatment | - | 2.29 | 1.10 | 0.00 | 3.00 |
| Control device condition | Treatment | - | 0.49 | 0.86 | 0.00 | 2.00 |
| Weather condition | Treatment | - | 1.84 | 0.44 | 0.00 | 2.00 |
| Improper turning involvement | Treatment | - | 0.17 | 0.38 | 0.00 | 1.00 |
| Alcohol or drug involvement | Treatment | - | 0.07 | 0.25 | 0.00 | 1.00 |
| Pedestrian involvement | Treatment | - | 0.03 | 0.18 | 0.00 | 1.00 |
| Cyclist involvement | Treatment | - | 0.02 | 0.13 | 0.00 | 1.00 |
| Motorcyclist involvement | Treatment | - | 0.03 | 0.18 | 0.00 | 1.00 |
| Population density | Confounder | People/km$^2$ | 2573.91 | 3272.42 | 0.02 | 59709.71 |
| Mean household income | Confounder | $ | 76516.76 | 42054.29 | 7709.00 | 437686.00 |
| Minority percentage | Confounder | % | 44.52 | 20.27 | 0.00 | 100.00 |
| Service sector job proportion | Confounder | % | 58.68 | 9.64 | 0.00 | 100.00 |
| Industrial sector job proportion | Confounder | % | 16.59 | 7.15 | 0.00 | 92.50 |
| Retail trade job proportion | Confounder | % | 14.65 | 5.19 | 0.00 | 100.00 |
| Transportation/warehousing job proportion | Confounder | % | 6.13 | 3.95 | 0.00 | 30.04 |
| Average road segment length | Confounder | m | 247.19 | 359.15 | 51.60 | 2913.03 |
| Intersection density | Confounder | num/km$^2$ | 36.00 | 23.46 | 0.02 | 168.88 |